\documentclass{article}

\usepackage{microtype}
\usepackage{graphicx}
\usepackage{subfigure}
\usepackage{booktabs} 
\usepackage{xcolor}     
\definecolor{champagne}{RGB}{247, 231, 206} 
\usepackage{colortbl}
\usepackage{color}
\usepackage{pifont}         
\usepackage{soul}           
\usepackage{xcolor}         
\usepackage{colortbl}       
\usepackage{multirow}       
\usepackage{hyperref}
\definecolor{darksalmon}{rgb}{0.91, 0.59, 0.48}
\definecolor{green(pigment)}{rgb}{0.0, 0.65, 0.31}
\usepackage{tcolorbox}

\usepackage[accepted]{icml2025}

\usepackage{amsmath}
\usepackage{amssymb}
\usepackage{mathtools}
\usepackage{amsthm}

\usepackage[capitalize,noabbrev]{cleveref}

\theoremstyle{plain}

\theoremstyle{definition}

\theoremstyle{remark}

\usepackage[textsize=tiny]{todonotes}
\usepackage{multirow}

\begin{document}

\twocolumn[
\icmltitle{PulseReddit: A Novel Reddit Dataset for Benchmarking MAS in High-Frequency Cryptocurrency Trading}




\icmlsetsymbol{equal}{*}

\begin{icmlauthorlist}
\icmlauthor{Qiuhan Han}{equal,tokyo}
\icmlauthor{Qian Wang}{equal,nus}
\icmlauthor{Atsushi Yoshikawa}{kanagawa}
\icmlauthor{Masayuki Yamamura}{tokyo}
\end{icmlauthorlist}

\icmlaffiliation{tokyo}{Institute of Science Tokyo}
\icmlaffiliation{nus}{National University of Singapore}
\icmlaffiliation{kanagawa}{Kanto Gakuin University}

\icmlcorrespondingauthor{Qiuhan Han}{han.q.ab@m.titech.ac.jp}

\icmlkeywords{Reddit Dataset, Cryptocurrency, Trade Strategy}

\vskip 0.3in
]
\printAffiliationsAndNotice{\icmlEqualContribution} 

\begin{abstract}
High-Frequency Trading (HFT) is pivotal in cryptocurrency markets, demanding rapid decision-making. Social media platforms like Reddit offer valuable, yet underexplored, information for such high-frequency, short-term trading. This paper introduces \textbf{PulseReddit}, a novel dataset that is the first to align large-scale Reddit discussion data with high-frequency cryptocurrency market statistics for short-term trading analysis. We conduct an extensive empirical study using Large Language Model (LLM)-based Multi-Agent Systems (MAS) to investigate the impact of social sentiment from PulseReddit on trading performance. Our experiments conclude that MAS augmented with PulseReddit data achieve superior trading outcomes compared to traditional baselines, particularly in bull markets, and demonstrate robust adaptability across different market regimes. Furthermore, our research provides conclusive insights into the performance-efficiency trade-offs of different LLMs, detailing significant considerations for practical model selection in HFT applications. PulseReddit and our findings establish a foundation for advanced MAS research in HFT, demonstrating the tangible benefits of integrating social media. The dataset is publicly available at \url{https://github.com/7huahua/RedditDataset}.
\end{abstract}

\section{Introduction}
Large Language Models (LLMs) have recently demonstrated considerable promise in financial decision-making \cite{liu2023fingpt, wu2023bloomberggpt}. 
The emergence of LLM-based Multi-Agent Systems (MAS) has significantly advanced complex task solving through role specialization and collaborative frameworks \cite{wang2024megaagent, wang2025agenttaxo}, while parallel breakthroughs in LLM reasoning capabilities \cite{wang2025assessing, chen2025judgelrm, wang2025limits} have enhanced analytical depth. These developments enable researchers to construct sophisticated agent ensembles—assigning distinct roles such as statistical analysts and trading agents—to develop more nuanced trading strategies \cite{li2024cryptotrade, yu2024finmem}. 
Despite these advancements, our review of the literature reveals a significant gap: current MAS predominantly focus on lower-frequency trading strategies, such as daily trading. Consequently, there is a notable scarcity of research applying these multi-agent systems to high-frequency trading (HFT) scenarios, such as 1-hour trading. This research void is critically exacerbated by the absence of publicly available datasets specifically tailored for MAS-based HFT research, particularly those integrating real-time social media signals with market data. This oversight is particularly striking given that HFT is a prevalent strategy actively employed by trading companies to identify and exploit fleeting market opportunities \cite{aldridge2013high, jones2013we, nahar2024market}.

To address these limitations, this paper introduces \textbf{PulseReddit}. This novel, large-scale dataset is meticulously curated from Reddit discussions concerning four major cryptocurrencies—BTC, ETH, DOGE, and SOL. It is synchronized with their corresponding high-frequency on-chain market data, covering intervals from 5-minutes to 4-hours. PulseReddit is specifically designed to facilitate research into the impact of social sentiment on cryptocurrency trading and to serve as a benchmark for developing advanced trading agents.

Furthermore, we conduct an extensive empirical study to evaluate PulseReddit's practical utility. This study employs Multi-Agent Systems (MAS) powered by leading Large Language Models such as GPT-4o, GPT-4o-mini, and DeepSeek-Chat. Our MAS framework uses specialized agents for market analysis, news analysis drawing on PulseReddit's social data, trading execution, and reflection for holistic decision-making. These MAS are benchmarked against traditional and learning-based strategies across diverse market conditions—bull, bear, and sideways—and various time granularities.

Our experiments lead to several key conclusions. First, we conclude that MAS augmented with social sentiment from PulseReddit achieve superior trading outcomes compared to traditional baselines, particularly in dynamic bull markets where they yielded up to 50\% higher returns. These systems also demonstrate robust adaptability across different market regimes. Second, our study concludes that integrating social signals from PulseReddit, while offering modest direct quantitative gains such as an average 5\% improvement in the Sharpe ratio, critically enables MAS to make more nuanced, risk-managed decisions. This is evidenced by qualitative case studies showing the synthesis of on-chain technicals with off-chain sentiment. Finally, our research provides a clear conclusion on LLM selection for HFT: while models like GPT-4o may yield peak performance, computationally lighter models such as GPT-4o-mini offer significant efficiency gains (e.g., 3-fold faster inference speeds) with only marginal performance trade-offs, making them viable for practical HFT deployment.

We make the following contributions:

\begin{enumerate}
    \item We provide the dataset, \textbf{PulseReddit}, that is the first to align large-scale Reddit discussion data with high-frequency cryptocurrency market statistics for short-term trading analysis.
    \item We conduct an extensive empirical study to evaluate PulseReddit's practical utility. We find conclusive demonstration that LLM-based Multi-Agent Systems, when leveraging social sentiment from PulseReddit, achieve superior performance compared to traditional baselines.
    \item We provide conclusive insights into the performance-efficiency trade-offs of different LLMs, guiding practical model selection for HFT applications.
\end{enumerate}



\section{Dataset Construction}
\subsection{Collection}

\begin{figure*}
    \centering
    \includegraphics[width=0.95\linewidth]{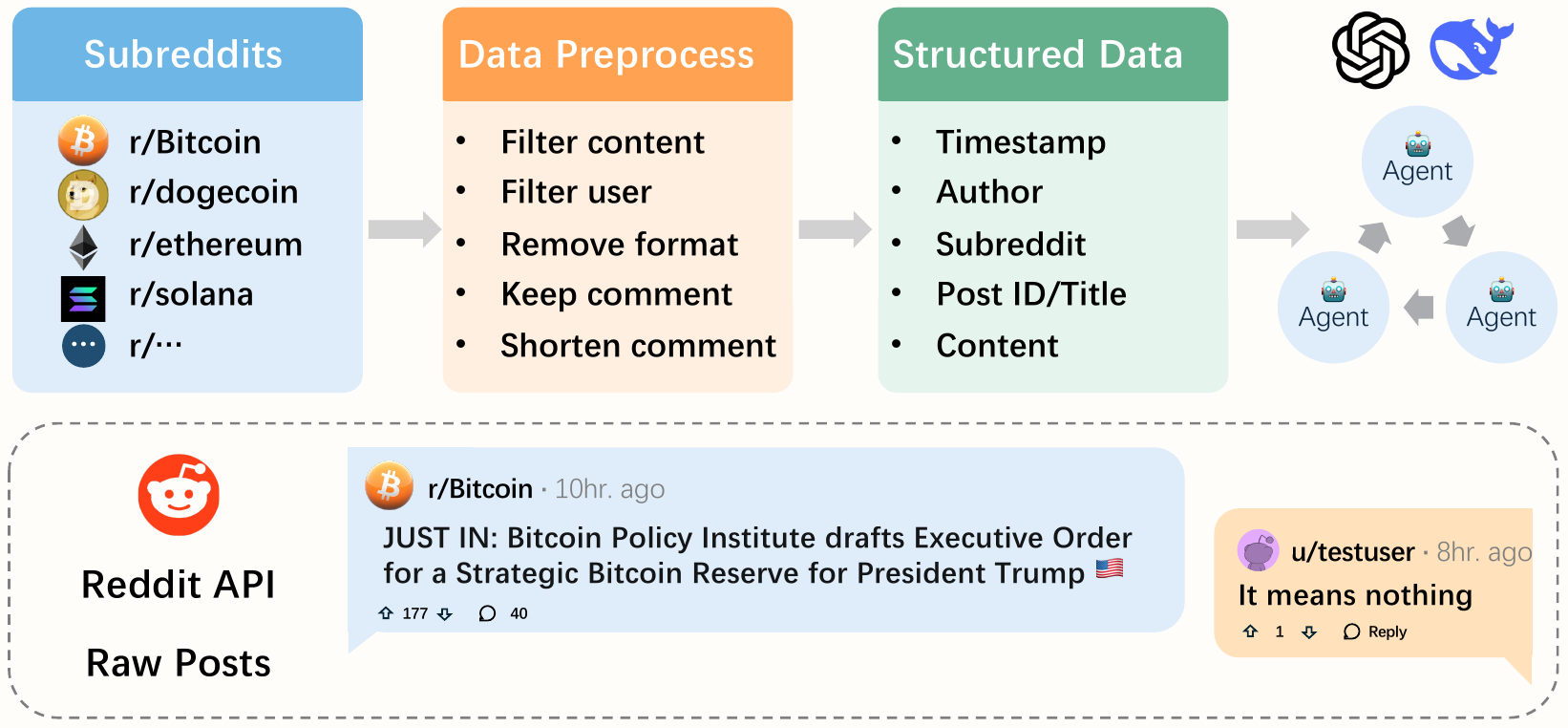}
    \caption{Overview of the PulseReddit dataset construction pipeline. Raw Reddit discussion data are collected via the Reddit API, undergo preprocessing and filtering, and are subsequently transformed into structured sentiment signals for downstream analysis in high-frequency multi-agent cryptocurrency trading.}

    \label{fig:reddit-pipeline}
\end{figure*}

Reddit is a social news aggregation and discussion platform composed of numerous topic-based forums called subreddits (denoted as `r/*`). Each subreddit forms a self-contained community centered around specific interests, where users publish posts and engage through threaded comments. Unlike traditional social media platforms, Reddit emphasizes topic-specific interaction and pseudo-anonymity, making it a rich source for understanding online opinion formation and temporal sentiment dynamics. We obey the ethics of Reddit as shown in Appendix \ref{data-ethics}.

In this study, we construct a cryptocurrency dataset by collecting posts and comments from six cryptocurrency-related subreddits: `r/Bitcoin`, `r/ethereum`, `r/dogecoin`, `r/solana`, `r/binance`, and `r/pepecoin`. These subreddits were selected based on their activity levels, market significance, and diversity in community composition. The data spans a one-year window from April 1st, 2024, to March 31st, 2025, covering a variety of market conditions including bull, bear, and sideways trends. 

In addition, a detailed view of the processing pipeline is shown in Figure~\ref{fig:reddit-pipeline}. Raw data is acquired via the Reddit API (or Pushshift API), followed by rigorous filtering to exclude deleted users, empty content, or link-containing entries. We also remove formatting characters and limit comment lengths between 10 and 100 words. The cleaned raw posts are then transformed into structured data, including timestamp, author ID, subreddit name, post title and content body. This structured format is ultimately used by the multi-agent system for downstream market trend inference and interaction modeling.

\subsubsection{Subreddit Overview}

\begin{table}[ht]
    \centering
    \caption{Reddit Subreddit Statistics for Six Cryptocurrencies (2024-04 to 2025-03).}
    \label{tab:subreddit_statistics}
    \resizebox{\linewidth}{!}{
    \begin{tabular}{l|c|c|c}
    \hline
    \textbf{Coin} & \textbf{Number of Posts} & \textbf{Avg. Words} & \textbf{Median Words} \\
    \hline
    BTC   & 28,798 & 68.16  & 36 \\
    DOGE  & 17,959 & 23.35  & 12 \\
    ETH   & 2,395  & 102.83 & 61 \\
    SOL   & 13,645 & 90.85  & 52 \\
    BNB   &   221  & 60.78  & 33 \\
    PEPE  & 9,597  & 42.59  & 23 \\
    \hline
    \end{tabular}
    }
\end{table}

Our dataset covers six major cryptocurrency subreddits, each representing a distinct coin and community focus. \texttt{r/Bitcoin} features extensive discussions on Bitcoin (BTC), including technical topics, market trends, and broader macroeconomic perspectives, making it the most information-rich and active among all. \texttt{r/ethereum} is oriented toward developments in Ethereum (ETH), with emphasis on decentralized finance, Layer-2 scaling, and protocol upgrades. \texttt{r/dogecoin} is shaped by a meme-driven investor base and highly responsive community sentiment, often reacting rapidly to viral events. \texttt{r/solana} centers on technical issues and growth of the Solana (SOL) ecosystem, while \texttt{r/binance} is focused on Binance Coin (BNB), covering exchange policies and trading features. \texttt{r/pepecoin} reflects the speculative dynamics of meme tokens, with frequent bursts of community activity. This selection enables systematic analysis of linguistic diversity, posting patterns, and sentiment shifts across fundamentally different coin communities.

Table~\ref{tab:subreddit_statistics} summarizes Reddit activity across six major cryptocurrencies between April 2024 and March 2025. Bitcoin (BTC) remains the most discussed coin, with nearly 29,000 unique posts, reflecting its sustained dominance in online community attention. Dogecoin (DOGE) also exhibits remarkably high engagement, with over 17,000 posts, highlighting its enduring social relevance and strong grassroots enthusiasm.

Other coins such as Ethereum (ETH), Solana (SOL), PEPE, and Binance Coin (BNB) show lower levels of Reddit discussion by comparison, though SOL and PEPE each experienced periods of heightened activity aligned with bullish market trends. Overall, the distribution of Reddit activity underscores the central role of BTC and DOGE in community-driven discourse within the crypto ecosystem.

\subsubsection{Data Filtering and Cleaning}

We adopt a systematic preprocessing pipeline, consistent with the DEBAGREEMENT dataset~\cite{pougue2021debagreement}. Specifically, we remove empty comments, deleted or hidden user content, and comments containing external URLs. All text is normalized by stripping special characters, and comments are truncated to a maximum of 100 words. To ensure focus on meaningful community interaction, we discard posts with fewer than the average number of replies and retain only first-level comments (i.e., direct replies to the original post). Additionally, we filter out comments shorter than 10 words or exceeding 100 words.

\subsection{Analysis}


\begin{figure}
    \centering
    \includegraphics[width=0.9\linewidth]{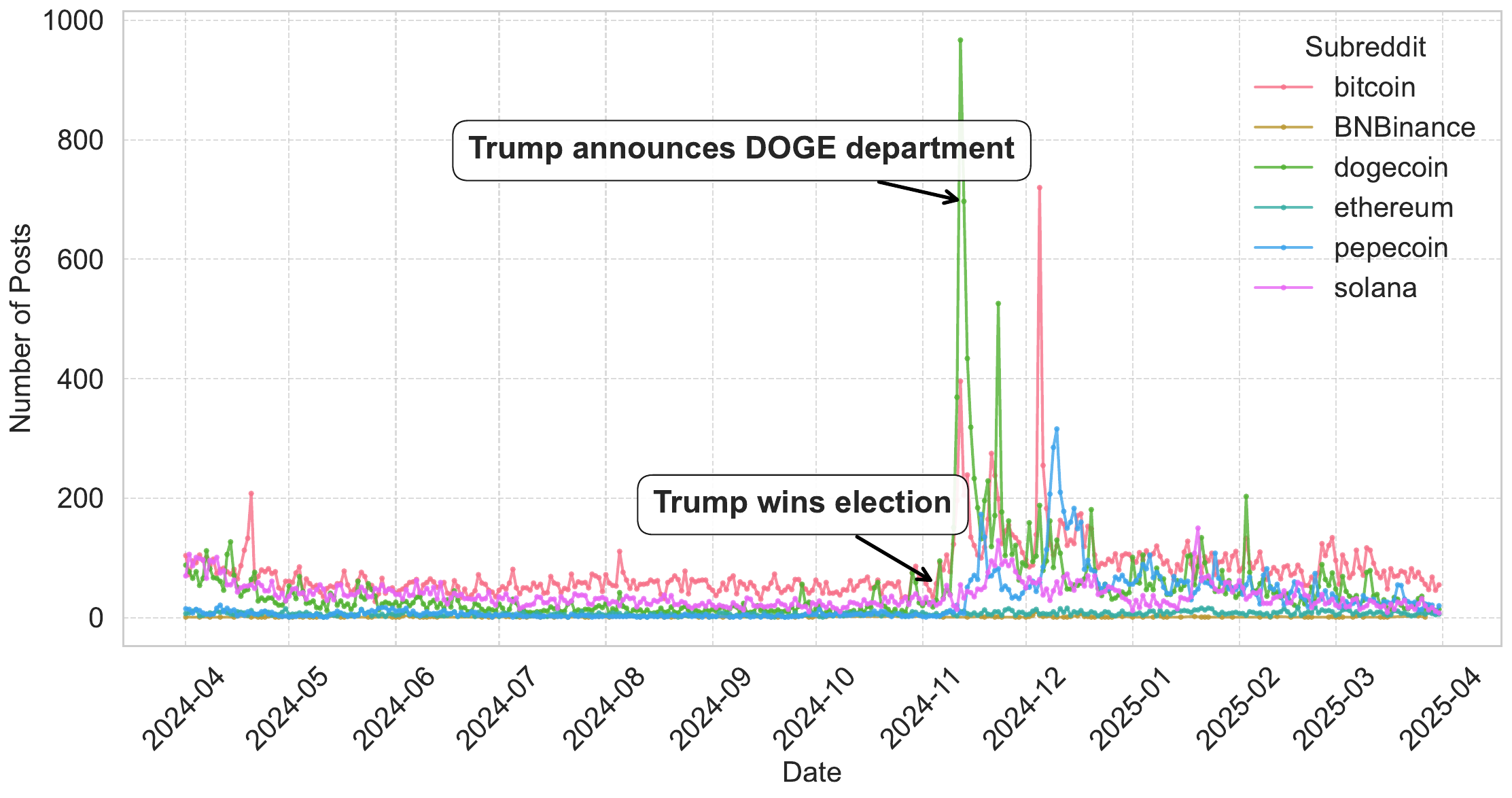}
    \caption{Daily post counts from six major cryptocurrency subreddits between April 2024 and March 2025. Sharp activity spikes correspond to two key events: the U.S. presidential election result and the announcement of a DOGE-focused government initiative.}
    \label{fig:post_count}
\end{figure}

To better understand the characteristics of the curated dataset, we analyzed the temporal dynamics of user interactions across the six selected subreddits over the period from April 2024 to March 2025. Figure~\ref{fig:post_count} illustrates the daily post volume for each subreddit. We observed that community activity generally remained stable at a moderate level throughout most of the year. However, two distinct surges in posting volume are evident across several subreddits. The first spike occurred in early November 2024, coinciding with the announcement that Donald Trump won the U.S. presidential election. The second, even more pronounced, occurred in late November 2024 when Trump publicly announced the formation of a DOGE-related federal department. This event triggered an unprecedented surge in `r/dogecoin`, where daily post volume temporarily exceeded 2,800. The ripple effect also caused notable increases in discussion activity across `r/bitcoin`, `r/ethereum`, and `r/pepecoin`, reflecting the interconnectedness of sentiment and news across different crypto communities.

These temporal patterns suggest that subreddit engagement is highly responsive to major geopolitical and crypto-specific announcements. Our dataset thus captures sentiment shifts across multiple communities aligned with real-world triggers, reinforcing its suitability for downstream tasks such as event-driven sentiment forecasting and market impact modeling.


\section{Experiments}

\subsection{Experimental Setup}

\textbf{Datasets.} Our experiments are conducted on four benchmark cryptocurrency datasets: BTC, ETH, DOGE, and SOL. Compared to previous works, these datasets not only contain standard price and trading volume data, but also incorporate richer multimodal and off-chain information. The off-chain news data is sourced from our curated PulseReddit dataset, which aggregates real-time Reddit discussions and sentiment signals for each coin. For the price data, we utilize historical records from Binance\footnote{\url{https://data.binance.vision/?prefix=data/spot/monthly/klines/}}, covering various time intervals to support high-frequency as well as longer-term trading simulations. For each coin, we collect aligned time-series of on-chain indicators (including open, close, high, low prices, and technical indicators) and off-chain news or sentiment information aggregated from relevant subreddit posts and comments. All datasets are pre-processed and synchronized to facilitate high-frequency trading simulations at multiple time granularities (5m, 15m, 1h, and 4h), ensuring the robustness and reproducibility of experimental results.


\textbf{Evaluation Metrics.} We initialize each trading strategy with a starting capital of one million US dollars. At the end of the trading simulation period, we assess performance using the \textit{Return} metric, which measures the overall profitability of the strategy. Specifically, Return is calculated as $\frac{w^{end}-w^{start}}{w^{start}}$, where $w^{start}$ and $w^{end}$ denote the initial and final net worth, respectively. 

\textbf{Baseline Strategies.} To benchmark the performance of MAS, we compare it against a comprehensive set of widely recognized baseline trading strategies. These strategies include the passive Buy and Hold benchmark, the theoretical Oracle upper bound, traditional rule-based technical indicators such as Simple Moving Average, Short-Long Moving Average, MACD, and Bollinger Bands, and a learning-based LSTM Prediction Strategy. All baseline strategies are evaluated across multiple time resolutions (5-minute, 15-minute, 1-hour and 4-hour) to assess performance under different temporal granularities. Detailed descriptions of each baseline strategy and their specific parameterizations are provided in Appendix~\ref{baselines}.

\textbf{MAS Setup.}
We adopt CryptoTrade~\cite{li2024cryptotrade} as our main multi-agent trading framework, designed to leverage the reasoning capabilities of Large Language Models (LLMs) for high-frequency cryptocurrency trading. CryptoTrade consists of four collaborative LLM-based agents—\textit{Market Analyst}, \textit{News Analyst}, \textit{Trading Agent}, and \textit{Reflection Agent}—each assigned a specialized role in analyzing and acting upon market signals. The system integrates both on-chain metrics (e.g., transaction volume, gas price, wallet activity) and off-chain signals (e.g., financial news, community sentiment from PulseReddit), simulating a decision-making pipeline that reflects real-world trading dynamics.

In each decision step (e.g., 5-minute intervals), the Market and News Analysts provide their respective summaries. The Trading Agent synthesizes these inputs into a continuous trading action ($[-1, 1]$ indicating proportional buy/sell volume). Post-episode, the Reflection Agent analyzes past performance to refine future strategies.

We implement this using GPT-4o, GPT-4o-mini and DeepSeek-chat (DeepSeek-V3), conducting zero-shot trials (no supervised fine-tuning) on BTC, ETH, SOL, and DOGE from our dataset across bull, bear, and sideways market phases. This configuration serves as a strong baseline to evaluate multi-agent coordination in realistic high-frequency trading conditions.




\textbf{Market Setup.} Table~\ref{tab:market_days_returns} summarizes the daily returns of four major cryptocurrencies—BTC, DOGE, ETH, and SOL—on selected days across three distinct market conditions: bull, bear, and sideways. These days are carefully chosen to represent typical behaviors under each condition. Bull market dates show significant positive returns across all assets, highlighting strong upward momentum. Bear market date are characterized by sharp negative returns, indicating sustained downward pressure. For the sideways market, we randomly selected three days with relatively neutral or mixed returns to reflect low directional trends. This setup provides a robust foundation for evaluating the trading strategies of our agent across varying market regimes.

\begin{table}[ht]
\centering
\caption{Daily Return (\%) of Major Coins on Selected Market Days.}
\label{tab:market_days_returns}
\resizebox{\linewidth}{!}{%
\begin{tabular}{l|c|c|c|c|c}
\hline
\textbf{Date} & \textbf{Market} & \textbf{BTC} & \textbf{DOGE} & \textbf{ETH} & \textbf{SOL} \\
\hline
\rowcolor{green!10} 2024-11-06 & Bull     & 8.937 & 15.57 & 12.35 & 11.97 \\
\rowcolor{green!10} 2024-11-10 & Bull     & 4.815 & 26.89 & 1.823 & 5.092\\
\rowcolor{green!10} 2024-11-11 & Bull     & 10.30 & 26.47 & 5.918 & 5.741\\
\rowcolor{red!10} 2024-04-13 & Bear     & -4.756 & -12.35 & -7.117 & -9.591\\
\rowcolor{red!10} 2025-02-24 & Bear     & -4.888 & -13.35 & -10.86 & -15.56\\
\rowcolor{red!10} 2025-03-03 & Bear     & -8.539 & -16.79 & -14.66 & -20.46\\
\hline
\end{tabular}
}
\end{table}

\textbf{Hardware.} All experiments are conducted on a single workstation equipped with an NVIDIA RTX 4090 GPU and 128 GB of system memory.

\textbf{Temperature.} We set the temperature to 0.7 for all experiments according to previous works \cite{li2024cryptotrade}.

\textbf{Prompts.} We provide the prompts used for the trading agents in Appendix~\ref{prompts}.

\subsection{Experimental Results}

Main results of baselines and MAS are shown in Table~\ref{tab:btc_results} to Table~\ref{tab:sol_results}. We have following findings:

In the bull market for BTC, MAS-based models, particularly GPT-4o and DeepSeek-Chat, consistently outperform traditional baselines across most time intervals. Table~\ref{tab:btc_results} presents the total return for these strategies. Notably, GPT-4o achieves the best result at the 5-minute interval (-0.90\%), while DeepSeek-Chat attains the highest returns at the 15-minute, 1-hour, and 4-hour intervals 

During bear markets on BTC, while performance declines across all strategies, MAS methods remain competitive. DeepSeek-Chat achieves the best results at 5 minutes (-13.48\%), while LSTM and SMA are able to achieve second-best performance at some frequencies. However, all strategies incur negative returns, highlighting the inherent difficulty of trading in adverse market conditions.

In sideways market regimes for BTC, GPT-4o-mini delivers the best performance at the 5-minute interval, while traditional baselines occasionally show competitive results at higher frequencies. Specifically, GPT-4o-mini achieved (-4.58\%) at 5 minutes, outperforming all other models. Traditional baselines such as MACD and SMA occasionally achieve competitive results at higher frequencies. Across most settings for BTC, MAS-based models either secure the highest or second-highest returns, demonstrating their robustness and adaptability.

These performance patterns, where MAS-based models generally show strong results, are largely consistent across other major cryptocurrencies like ETH, DOGE, and SOL. As detailed in the supplementary tables (Table~\ref{tab:doge_results}, Table~\ref{tab:eth_results}, and Table~\ref{tab:sol_results}), MAS-based models, especially GPT-4o and DeepSeek-Chat, also generally surpass traditional baselines in bull and sideways market conditions, while retaining competitive performance even in bear markets. The performance ranking and relative advantages between strategies remain largely consistent, underscoring the generalizability and robustness of MAS-based approaches across different coins and market scenarios.

\begin{table*}[ht]
\centering
\caption{Total Return (\%) of baseline and MAS strategies on BTC across market regimes and time frequencies. Best and second-best results in each market-frequency block are \textbf{bolded} and \underline{underlined}, respectively.}

\label{tab:btc_results}
\resizebox{\textwidth}{!}{%
\begin{tabular}{l|l|cccc|cccc|cccc}
\hline
\textbf{Dataset} & \textbf{Strategy} & \multicolumn{4}{c|}{\textbf{Bull Market}} & \multicolumn{4}{c|}{\textbf{Bear Market}} & \multicolumn{4}{c|}{\textbf{Sideways Market}} \\
\cline{3-14}
& & 5m & 15m & 1h & 4h & 5m & 15m & 1h & 4h & 5m & 15m & 1h & 4h \\
\hline

\multirow{7}{*}{BTC} 
& LSTM & -9.09 & -4.86 & -1.86 & -0.68 & -15.55 & \underline{-7.03} & -4.50 & \underline{-3.03} & \underline{-9.09} & -4.86 & \underline{-1.86} & \underline{-0.68} \\
& SMA  & -10.00 & \underline{0.38} & \underline{3.28} & 3.41 & -14.62 & \textbf{-6.62} & \textbf{-3.04} & \textbf{-3.00 }& -14.07 & -4.38 & \textbf{-1.18} & -0.75 \\
& MACD & -8.26 & -1.65 & 0.78 & 2.07 & -14.45 & -7.85 & -5.12 & -4.48 & -10.71 & \textbf{-3.90} & -2.32 & \textbf{-0.07} \\

\cline{2-14}
& gpt-4o        & \textbf{-0.90} & -1.10 & 2.86 & \underline{3.42} & -13.99 & -8.32 & -5.00 & -4.58 & -10.84 & -4.34 & -2.72 & -1.77 \\
& gpt-4o-mini   & -6.40 & -1.18 & 1.80 & 1.79 & \underline{-13.63} & -7.58 & \underline{-4.14} & -3.65 & \textbf{-4.58}  & \underline{-4.06} & -2.62 & -1.65 \\
& deepseek-chat & \underline{-3.26} & \textbf{3.14}  & \textbf{5.10 }& \textbf{5.40} & \textbf{-13.48} & -8.53 & -5.11 & -4.04 & -10.91 & -5.32 & -2.03 & -1.75 \\

\hline
\end{tabular}
}
\end{table*}

\begin{table*}[ht]
\centering
\caption{Total Return (\%) of baseline and MAS strategies on DOGE across market regimes and time frequencies. Best and second-best results in each market-frequency block are \textbf{bolded} and \underline{underlined}, respectively.}
\label{tab:doge_results}
\resizebox{\textwidth}{!}{%
\begin{tabular}{l|l|cccc|cccc|cccc}
\hline
\textbf{Dataset} & \textbf{Strategy} & \multicolumn{4}{c|}{\textbf{Bull Market}} & \multicolumn{4}{c|}{\textbf{Bear Market}} & \multicolumn{4}{c|}{\textbf{Sideways Market}} \\
\cline{3-14}
& & 5m & 15m & 1h & 4h & 5m & 15m & 1h & 4h & 5m & 15m & 1h & 4h \\
\hline
\multirow{7}{*}{DOGE} 
& LSTM & -11.00 & -6.84 & -3.37 & -2.08 & -19.68 & -13.77 & -11.04 & -7.08 & -11.00 & -6.84 & -3.37 & -2.08 \\{}
& SMA & -24.08 & -4.18 & 3.99 & -0.42 & -22.49 & -12.74 & \underline{-6.88} & \underline{-6.87} & -14.77 & -4.627 & \underline{-1.807} & -2.163 \\
& MACD & -4.67 & -5.26 & 9.28 & \underline{4.17} & \underline{-16.12} & \underline{-11.89} & -15.64 & -13.80 & \underline{-8.491}  & -6.466 & -2.961 & -3.954 \\

\cline{2-14}
& gpt-4o & \underline{-1.03} & \underline{4.21} & 9.80 & 3.31 & -18.56 & -13.25 & -10.17 & -8.52 & -8.82 & \underline{-5.45} & -3.20 & -3.90 \\
& gpt-4o-mini & -2.15 & 3.47 & \underline{9.96} & 1.57 & -21.01 & -13.20 & -10.74 & -8.59 & -9.00 & \underline{-5.45} & -3.33 & -4.25 \\
& deepseek-chat & \textbf{0.33} & \textbf{9.10} & \textbf{16.00} & \textbf{7.21} & \textbf{-18.56} & \textbf{-12.74} & \textbf{-6.88} & \textbf{-6.87} & \textbf{-8.491} & \textbf{-4.627} & \textbf{-1.807} & \textbf{-2.08} \\
\hline
\end{tabular}
}
\end{table*}

\begin{table*}[ht]
\centering
\caption{Total Return (\%) of baseline and MAS strategies on ETH across market regimes and time frequencies. Best and second-best results in each market-frequency block are \textbf{bolded} and \underline{underlined}, respectively.}
\label{tab:eth_results}
\resizebox{\textwidth}{!}{%
\begin{tabular}{l|l|cccc|cccc|cccc}
\hline
\textbf{Dataset} & \textbf{Strategy} & \multicolumn{4}{c|}{\textbf{Bull Market}} & \multicolumn{4}{c|}{\textbf{Bear Market}} & \multicolumn{4}{c|}{\textbf{Sideways Market}} \\
\cline{3-14}
& & 5m & 15m & 1h & 4h & 5m & 15m & 1h & 4h & 5m & 15m & 1h & 4h \\
\hline
\multirow{7}{*}{ETH}
& LSTM & -10.07 & -4.20 & -1.95 & -0.93 & -17.93 & -8.30 & -8.14 & \underline{-5.44} & -10.07 & -4.20 & -1.95 & \underline{-0.93 }\\
& SMA & -19.24 & -5.96 &\underline{ 0.67} & -2.37 & -22.45 & -12.73 & -7.93 & -6.28 & -10.94 & \underline{-3.94} & \underline{-1.23} & -0.97 \\
& MACD & -9.45 & -6.52 & \textbf{1.61} & -1.19 & \textbf{-14.03} & -11.59 & -13.79 & -12.46 & \underline{-8.97}  & -4.11 & -2.44 & -1.05 \\

\cline{2-14}
& gpt-4o & \underline{-5.91} & \underline{-0.80} & 0.24 & \textbf{1.52} & -15.25 & \textbf{-10.63} & -7.72 & -5.71 & -9.30 & -4.90 & -1.89 & -1.99 \\
& gpt-4o-mini & -6.45 & -1.23 & 0.40 & \underline{1.38} & -16.14 & \underline{-10.63} & \underline{-7.63} & \textbf{-4.83} & -10.01 & -4.28 & \textbf{-1.19} & -2.02 \\
& deepseek-chat & \textbf{-5.81} & \textbf{-0.76} & 0.50 & \underline{1.38} & \underline{-14.28} & -11.29 & \textbf{-7.63} & -7.87 & \textbf{-8.95} & \textbf{-3.94} & -1.89 & \textbf{-0.93} \\

\hline
\end{tabular}
}
\end{table*}

\begin{table*}[ht]
\centering
\caption{Total Return (\%) of baseline and MAS strategies on SOL across market regimes and time frequencies. Best and second-best results in each market-frequency block are \textbf{bolded} and \underline{underlined}, respectively.}
\label{tab:sol_results}
\resizebox{\textwidth}{!}{%
\begin{tabular}{l|l|cccc|cccc|cccc}
\hline
\textbf{Dataset} & \textbf{Strategy} & \multicolumn{4}{c|}{\textbf{Bull Market}} & \multicolumn{4}{c|}{\textbf{Bear Market}} & \multicolumn{4}{c|}{\textbf{Sideways Market}} \\
\cline{3-14}
& & 5m & 15m & 1h & 4h & 5m & 15m & 1h & 4h & 5m & 15m & 1h & 4h \\
\hline
\multirow{7}{*}{SOL} 
& LSTM & -11.95 & -4.66 & -4.64 & -2.41 & -17.93 & -8.30 & -8.14 & \underline{-5.44} & -11.95 & -4.66 & -4.64 & -2.41 \\
& SMA & -22.19 & -9.54 & -0.13 & 0.79 & -21.26 & -14.08 & \underline{-7.29} & -8.19 & -11.88 & -3.94 & \underline{-1.54} & \underline{-1.67} \\
& MACD & \textbf{-3.37} & -4.88 & -1.11 & 2.46 & -19.55 & -16.24 & -19.26 & -17.08 & \textbf{-10.13} & -4.50 & -3.86 &  \textbf{0.23} \\

\cline{2-14}
& gpt-4o & \underline{-3.88} & \underline{-1.51} & \textbf{3.15} & \underline{2.92} & -18.29 & -13.59 & -11.58 & -7.75 & -11.21 & -3.45 & -3.74 & -1.97 \\
& gpt-4o-mini & -5.22 & -1.52 & 1.07 & 1.68 & -19.51 & -13.28 & -10.88 & -8.85 & -11.52 & \textbf{-3.09} & -4.35 & -0.74 \\
& deepseek-chat & -4.80 & \textbf{0.85} & \underline{2.28} & \textbf{3.06} & \textbf{-16.72} & \textbf{-8.30} & \textbf{-7.29} & \textbf{-5.44} & \underline{-10.89} & \underline{-3.45} & \textbf{-1.54} & -2.41 \\

\hline
\end{tabular}
}
\end{table*}

The computational efficiency varies significantly among the LLM agents, with GPT-4o-mini consistently achieving the lowest execution time across all coins and time frequencies. As shown in Table~\ref{tab:runtime_comparison_minutes}, 4o-mini often completes simulation runs in less than half the time required by gpt-4o and less than a third of that required by deepseek-chat at the 5-minute trading interval. For example, the average execution time for BTC at 5m is 102.35 minutes for 4o-mini, compared to 134.18 minutes for gpt-4o and 249.96 minutes for deepseek-chat. This efficiency gap persists across all coins.

As trading frequency decreases (i.e., from 5m to 4h intervals), the execution time for all models drops significantly, reflecting the reduced number of required agent decisions. This trend is evident across all models in Table~\ref{tab:runtime_comparison_minutes}. Nevertheless, 4o-mini maintains the fastest inference speed under every setting.

Considering these efficiency results, 4o-mini is best suited for rapid experimentation and large-batch simulation scenarios where computational resources or turnaround time are critical constraints. GPT-4o provides a reasonable trade-off between model strength and efficiency. In contrast, deepseek-chat is substantially slower in all cases and may only be suitable for low-frequency simulations or qualitative case studies due to its high latency. This efficiency analysis should inform the selection of model backends when designing automated trading systems or large-scale agent evaluations.

\begin{table}[ht]
\centering
\caption{Average Execution Time (minutes) for Each Model and Coin.}
\label{tab:runtime_comparison_minutes}
\resizebox{\linewidth}{!}{%
\begin{tabular}{l|l|r|r|r|r}
\hline
\textbf{Model} & \textbf{Coin} & \textbf{5m} & \textbf{15m} & \textbf{1h} & \textbf{4h} \\
\hline
gpt-4o         & BTC  & 134.18 & 50.35 & 13.27 & 2.45 \\
               & DOGE & 122.63 & 21.89 & 6.64  & 1.46 \\
               & ETH  & 126.20 & 29.22 & 7.25  & 1.76 \\
               & SOL  & 124.44 & 32.51 & 8.88  & 2.12 \\
\hline
4o-mini        & BTC  & 102.35 & 44.50 & 5.82  & 1.42 \\
               & DOGE & 98.05  & 44.50 & 6.06  & 1.53 \\
               & ETH  & 85.37  & 39.57 & 8.30  & 2.07 \\
               & SOL  & 91.68  & 40.61 & 7.92  & 2.17 \\
\hline
deepseek-chat  & BTC  & 249.96 & 93.51 & 19.87 & 4.98 \\
               & DOGE & 247.92 & 84.73 & 16.83 & 4.28 \\
               & ETH  & 212.94 & 103.80 & 14.17 & 3.37 \\
               & SOL  & 204.95 & 79.92 & 14.22 & 3.47 \\
\hline
\end{tabular}
}
\end{table}

\subsection{Ablation study}
To evaluate the impact of the PulseReddit dataset, we conducted an ablation study comparing agent performance with and without Reddit-based news data. The results are shown in Table~\ref{tab:ablation_reddit_vertical}. We have following findings:

Across both bull and bear market conditions, the inclusion of Reddit information leads to a slight but consistent improvement in both total return and Sharpe ratio. Specifically, in bull markets, the total return improves from -5.23\% (without Reddit) to -4.25\% (with Reddit), and the Sharpe ratio increases from -0.11 to -0.08. Similarly, in bear markets, the addition of Reddit data results in a marginal reduction in losses, with the total return improving from -11.5\% to -11.4\%, and the Sharpe ratio rising from -0.25 to -0.24.

The agent is able to extract useful market sentiment and event signals from Reddit discussions, particularly in periods of heightened market activity, although the overall effect size is modest. The performance improvement is more pronounced in bull markets, possibly due to greater news-driven price movements during such periods.


Overall, incorporating Reddit as an auxiliary information source contributes to a more robust agent, albeit with marginal quantitative gains under the tested settings. This underscores the potential value of further enhancing the quality or granularity of social sentiment data to achieve stronger downstream impact on trading strategies.

\begin{table}[ht]
\centering
\caption{Effect of Reddit Information (with vs. without) across Market Conditions.}
\label{tab:ablation_reddit_vertical}
\begin{tabular}{l|l|cc}
\hline
\textbf{Market} & \textbf{Metric} & \textbf{w/o Reddit} & \textbf{w/ Reddit} \\
\hline
Bull  & Total Return   & -5.23  & \textbf{-4.25} \\
      & Sharpe Ratio   & -0.11  & \textbf{-0.08} \\
\hline
Bear  & Total Return   & -11.5  & \textbf{-11.4} \\
      & Sharpe Ratio   & -0.25  & \textbf{-0.24} \\
\hline
\end{tabular}
\end{table}

\subsection{Case Study}

To demonstrate the practical decision-making capabilities of our multi-agent system in real market scenarios, we present a detailed case study of the DOGE FOMO event on 2024-11-10, as shwon in Table \ref{tab:doge_fomo_case}. This case illustrates how the system integrates conflicting on-chain and off-chain signals to make prudent trading decisions during periods of extreme market sentiment. We have following findings:

Reddit headlines overwhelmingly reflected a strong retail "fear of missing out," with the community predicting imminent price surges and new highs. Table~\ref{tab:doge_fomo_case} shows the step-by-step reasoning process during this period of extreme FOMO in the Dogecoin market. The system first receives a series of highly bullish Reddit headlines, reflecting widespread retail optimism and speculative sentiment.

Despite these strong off-chain signals, the technical analysis agent identifies bearish on-chain indicators and warns of potential reversal risks. While the news analyst agent detected this heightened optimism, the market analyst agent identified a persistent bearish technical indicator (MACD: sell) and cautioned about the risk of a blow-off top and subsequent reversal.

In response, the decision-making agent opted for a partial reduction in exposure, executing a light sell action (-0.3), while maintaining a core position. This conservative approach was justified in hindsight: although DOGE briefly spiked following the surge in retail enthusiasm, the price quickly retraced, validating the agent's cautious stance.


\begin{table*}[ht]
\centering
\caption{Case Study: Agent Reasoning and Decisions During DOGE FOMO Event (2024-11-10).}
\label{tab:doge_fomo_case}
\begin{tabular}{p{3.2cm} | p{11.0cm}}
\hline
\textbf{Agent / Step} & \textbf{Content} \\
\hline
News Analyst &
\textbf{Input:} \newline
You are required to analyze the following news articles: \newline
1. TO THE MOON BABY! .23 before midnight! \newline
2. 25cents before daybreak!! LFG!!!! \newline
3. DOGE momentum continues as retail crowd jumps in. \newline
\textbf{Output:} \newline
The news headlines are extremely bullish, with strong retail sentiment and FOMO. Multiple posts are anticipating DOGE breaking key price levels (0.23, 0.25) in the near term. Community is highly optimistic, typical of FOMO-driven spikes. \\
\hline
Market Analyst &
\textbf{Input:} \newline
Open price: 0.220, MACD: sell \newline
\textbf{Output:} \newline
Despite the bullish news and community enthusiasm, the on-chain indicator MACD remains bearish. There is a risk of a blow-off top followed by sharp correction, as previous FOMO-driven rallies often result in quick reversals. \\
\hline
Decision Maker &
\textbf{Input:} \newline
(news and technical analysis above) \newline
\textbf{Output:} \newline
\textbf{Action: -0.3} (Light sell / partial reduction). Reason: Maintain core position but reduce exposure given strong FOMO and lack of confirmation from technical indicators. \\
\hline
Reflection Agent &
\textbf{Input:} \newline
Market move and agent's prior actions \newline
\textbf{Output:} \newline
The reduction in position limited downside during the reversal after the short-lived rally. Overexposure could have resulted in significant losses. The decision to stay cautious and not chase the FOMO appears justified in hindsight. \\
\hline
\end{tabular}
\end{table*}

\section{Related Work}

\textbf{LLM-based Financial Applications.} Recent progress in Large Language Models (LLMs) has significantly impacted a wide range of financial decision-making processes. General-purpose financial LLMs such as FinGPT~\cite{liu2023fingpt}, BloombergGPT~\cite{wu2023bloomberggpt}, and FinMA~\cite{xie2023pixiu} primarily target traditional equity and commodity markets, focusing on information extraction and forecasting from off-chain textual sources. Building on this foundation, several agent-based frameworks—including FinAgent~\cite{zhang2024multimodal}, FinMem~\cite{yu2024finmem}, and Sociodojo~\cite{cheng2024sociodojo}—have leveraged LLMs to automate and enhance stock market trading strategies, portfolio management, and event-driven decision-making. Notably, FinMem systematically compares different LLM backbones for financial tasks, highlighting the sensitivity of performance to underlying language model architectures. Within the cryptocurrency domain, CryptoTrade~\cite{li2024reflective} introduces a reflective multi-agent framework, enabling adaptive, LLM-based trading agents to interact with highly volatile crypto markets.

\textbf{Cryptocurrency Datasets.}
A growing body of research has contributed public cryptocurrency datasets to advance various tasks such as price forecasting, transaction graph modeling, and account classification~\citep{lin2020modeling, lin2022ethereum, wang2022tsgn, li2024cryptotrade, huang2022ethereum, zhang2024live,luo2024multi}. Additional efforts address specialized problems like phishing account detection~\citep{chen2020phishing, li2022ttagn, wang2023ex}. However, most existing datasets are built exclusively on on-chain transaction records, resulting in limited node features and generally lacking high-quality textual or social context. In particular, there is a notable absence of datasets centered on cryptocurrency-related discussions, especially those capturing real-time or short-term market sentiment from social media. Moreover, prior datasets rarely provide the time granularity, off-chain integration, or scale required for rigorous evaluation of high-frequency trading systems. To highlight the novelty of our PulseReddit dataset in addressing these gaps—particularly its focus on Reddit data and its applicability to short-term trading research—we provide a detailed comparison with previous works in Table~\ref{tab:dataset_comparison}.

\begin{table*}[!h]
  \centering
  \caption{Comparison of Cryptocurrency Datasets Highlighting Data Sources and Suitability for Short-Term Trading.}
  \label{tab:dataset_comparison}
  \resizebox{\textwidth}{!}{%
    \begin{tabular}{l|c|c|c|c}
    \toprule
    \rowcolor{champagne}
    \textbf{Dataset} & \textbf{Social Media Data} & \textbf{Includes Reddit Data} & \textbf{For Short-Term Trading} & \textbf{Publicly Available} \\
    \midrule
    \rowcolor{gray!10}
    Physica A\cite{lin2022ethereum} & \textcolor{darksalmon}{\ding{55}} & \textcolor{darksalmon}{\ding{55}} & \textcolor{darksalmon}{\ding{55}} & \textcolor{green(pigment)}{\ding{51}} \\
    EH-GCN \cite{huang2022ethereum} & \textcolor{darksalmon}{\ding{55}} & \textcolor{darksalmon}{\ding{55}} & \textcolor{darksalmon}{\ding{55}} & \textcolor{darksalmon}{\ding{55}} \\
    \rowcolor{gray!10}
    TTAGN \cite{li2022ttagn} & \textcolor{darksalmon}{\ding{55}} & \textcolor{darksalmon}{\ding{55}} & \textcolor{darksalmon}{\ding{55}} & \textcolor{darksalmon}{\ding{55}} \\
    Chartlist \cite{shamsi2022chartalist} & \textcolor{darksalmon}{\ding{55}} & \textcolor{darksalmon}{\ding{55}} & \textcolor{darksalmon}{\ding{55}} & \textcolor{green(pigment)}{\ding{51}} \\
    \rowcolor{gray!10}
    EX-Graph \cite{wang2023ex} & \textcolor{green(pigment)}{\ding{51}} & \textcolor{darksalmon}{\ding{55}} & \textcolor{darksalmon}{\ding{55}} & \textcolor{green(pigment)}{\ding{51}} \\
    CryptoTrade \cite{li2024cryptotrade} & \textcolor{green(pigment)}{\ding{51}} & \textcolor{darksalmon}{\ding{55}} & \textcolor{green(pigment)}{\ding{51}} & \textcolor{green(pigment)}{\ding{51}} \\
    \midrule
    \rowcolor{gray!10}
    \textbf{PulseReddit (Ours)} & \textcolor{green(pigment)}{\ding{51}} & \textcolor{green(pigment)}{\ding{51}} & \textcolor{green(pigment)}{\ding{51}} & \textcolor{green(pigment)}{\ding{51}} \\
    \bottomrule
    \end{tabular}
    }
\end{table*}

\section{Conclusion}

In this paper, we introduced PulseReddit, the first large-scale dataset synchronizing Reddit discussions with high-frequency cryptocurrency market statistics for short-term trading research. Through extensive empirical evaluation using LLM-based Multi-Agent Systems across GPT-4o, GPT-4o-mini, and DeepSeek-Chat, we demonstrated that MAS augmented with PulseReddit data consistently outperform traditional baselines, particularly in bull markets. Our ablation study shows that incorporating Reddit data leads to consistent improvements in both total return and Sharpe ratio across market conditions, with performance gains more pronounced during periods of heightened market activity. The case study of the DOGE FOMO event illustrates how our system synthesizes conflicting on-chain and off-chain signals to make prudent risk-managed decisions. Additionally, our efficiency analysis reveals significant performance-efficiency trade-offs among LLMs, with GPT-4o-mini achieving substantially faster inference speeds while maintaining competitive performance, providing practical guidance for HFT deployment.

\subsection*{Limitations and Broader Impacts}
Our study focuses on four major cryptocurrencies and Reddit as the primary social media source, which may limit generalizability to other assets or platforms. The dataset reflects inherent biases of Reddit communities, and the modest quantitative improvements from social signals suggest that on-chain indicators remain primary drivers for short-term trading. These trading strategies are intended for academic research only and should not be used as investment advice. The use of social media data in finance raises ethical considerations regarding data privacy and potential market manipulation that require careful consideration in future applications.

\hspace{1em}

\textbf{Acknowledgements.} This research was funded by JST SPRING, Japan, Grant Number JPMJSP2180.

\hspace{1em}

\nocite{langley00}

\bibliography{main}

\begin{thebibliography}{31}
\providecommand{\natexlab}[1]{#1}
\providecommand{\url}[1]{\texttt{#1}}
\expandafter\ifx\csname urlstyle\endcsname\relax
  \providecommand{\doi}[1]{doi: #1}\else
  \providecommand{\doi}{doi: \begingroup \urlstyle{rm}\Url}\fi

\bibitem[Aldridge(2013)]{aldridge2013high}
Aldridge, I.
\newblock \emph{High-frequency trading: a practical guide to algorithmic strategies and trading systems}.
\newblock John Wiley \& Sons, 2013.

\bibitem[Chen et~al.(2020)Chen, Peng, Liu, Li, Xie, and Zheng]{chen2020phishing}
Chen, L., Peng, J., Liu, Y., Li, J., Xie, F., and Zheng, Z.
\newblock Phishing scams detection in ethereum transaction network.
\newblock \emph{ACM Transactions on Internet Technology (TOIT)}, 21\penalty0 (1):\penalty0 1--16, 2020.

\bibitem[Chen et~al.(2025)Chen, Hu, Zou, Wu, Wang, Hooi, and He]{chen2025judgelrm}
Chen, N., Hu, Z., Zou, Q., Wu, J., Wang, Q., Hooi, B., and He, B.
\newblock Judgelrm: Large reasoning models as a judge.
\newblock \emph{arXiv preprint arXiv:2504.00050}, 2025.

\bibitem[Cheng \& Chin(2024)Cheng and Chin]{cheng2024sociodojo}
Cheng, J. and Chin, P.
\newblock Sociodojo: Building lifelong analytical agents with real-world text and time series.
\newblock In \emph{The Twelfth International Conference on Learning Representations}, 2024.
\newblock URL \url{https://openreview.net/forum?id=s9z0HzWJJp}.

\bibitem[Ferdiansyah et~al.(2019)Ferdiansyah, Othman, Radzi, Stiawan, Sazaki, and Ependi]{ferdiansyah2019lstm}
Ferdiansyah, F., Othman, S.~H., Radzi, R. Z. R.~M., Stiawan, D., Sazaki, Y., and Ependi, U.
\newblock A lstm-method for bitcoin price prediction: A case study yahoo finance stock market.
\newblock In \emph{2019 international conference on electrical engineering and computer science (ICECOS)}, pp.\  206--210. IEEE, 2019.

\bibitem[Gencay(1996)]{gencay1996non}
Gencay, R.
\newblock Non-linear prediction of security returns with moving average rules.
\newblock \emph{Journal of Forecasting}, 15\penalty0 (3):\penalty0 165--174, 1996.

\bibitem[Huang et~al.(2022)Huang, Lin, and Wu]{huang2022ethereum}
Huang, T., Lin, D., and Wu, J.
\newblock Ethereum account classification based on graph convolutional network.
\newblock \emph{IEEE Transactions on Circuits and Systems II: Express Briefs}, 69\penalty0 (5):\penalty0 2528--2532, 2022.

\bibitem[Jones(2013)]{jones2013we}
Jones, C.~M.
\newblock What do we know about high-frequency trading?
\newblock \emph{Columbia Business School Research Paper}, \penalty0 (13-11), 2013.

\bibitem[Langley(2000)]{langley00}
Langley, P.
\newblock Crafting papers on machine learning.
\newblock In Langley, P. (ed.), \emph{Proceedings of the 17th International Conference on Machine Learning (ICML 2000)}, pp.\  1207--1216, Stanford, CA, 2000. Morgan Kaufmann.

\bibitem[Li et~al.(2022)Li, Gou, Liu, Hou, Li, and Xiong]{li2022ttagn}
Li, S., Gou, G., Liu, C., Hou, C., Li, Z., and Xiong, G.
\newblock Ttagn: Temporal transaction aggregation graph network for ethereum phishing scams detection.
\newblock In \emph{Proceedings of the ACM Web Conference 2022}, pp.\  661--669, 2022.

\bibitem[Li et~al.(2024{\natexlab{a}})Li, Luo, Wang, Chen, Liu, and He]{li2024cryptotrade}
Li, Y., Luo, B., Wang, Q., Chen, N., Liu, X., and He, B.
\newblock Cryptotrade: A reflective llm-based agent to guide zero-shot cryptocurrency trading.
\newblock In \emph{Proceedings of the 2024 Conference on Empirical Methods in Natural Language Processing}, pp.\  1094--1106, 2024{\natexlab{a}}.

\bibitem[Li et~al.(2024{\natexlab{b}})Li, Luo, Wang, Chen, Liu, and He]{li2024reflective}
Li, Y., Luo, B., Wang, Q., Chen, N., Liu, X., and He, B.
\newblock A reflective llm-based agent to guide zero-shot cryptocurrency trading.
\newblock \emph{arXiv preprint arXiv:2407.09546}, 2024{\natexlab{b}}.

\bibitem[Lin et~al.(2020)Lin, Wu, Yuan, and Zheng]{lin2020modeling}
Lin, D., Wu, J., Yuan, Q., and Zheng, Z.
\newblock Modeling and understanding ethereum transaction records via a complex network approach.
\newblock \emph{IEEE Transactions on Circuits and Systems II: Express Briefs}, 67\penalty0 (11):\penalty0 2737--2741, 2020.

\bibitem[Lin et~al.(2022)Lin, Wu, Xuan, and Chi]{lin2022ethereum}
Lin, D., Wu, J., Xuan, Q., and Chi, K.~T.
\newblock Ethereum transaction tracking: Inferring evolution of transaction networks via link prediction.
\newblock \emph{Physica A: Statistical Mechanics and its Applications}, 600:\penalty0 127504, 2022.

\bibitem[Liu et~al.(2023)Liu, Wang, and Zha]{liu2023fingpt}
Liu, X.-Y., Wang, G., and Zha, D.
\newblock Fingpt: Democratizing internet-scale data for financial large language models.
\newblock \emph{arXiv preprint arXiv:2307.10485}, 2023.

\bibitem[Luo et~al.(2024)Luo, Zhang, Wang, and He]{luo2024multi}
Luo, B., Zhang, Z., Wang, Q., and He, B.
\newblock Multi-chain graphs of graphs: A new approach to analyzing blockchain datasets.
\newblock \emph{Advances in Neural Information Processing Systems}, 37:\penalty0 28490--28514, 2024.

\bibitem[Nahar et~al.(2024)Nahar, Nishat, Shoaib, and Hossain]{nahar2024market}
Nahar, J., Nishat, N., Shoaib, A., and Hossain, Q.
\newblock Market efficiency and stability in the era of high-frequency trading: A comprehensive review.
\newblock \emph{International Journal of Business and Economics}, 1\penalty0 (3):\penalty0 1--13, 2024.

\bibitem[Pougu{\'e}-Biyong et~al.(2021)Pougu{\'e}-Biyong, Semenova, Matton, Han, Kim, Lambiotte, and Farmer]{pougue2021debagreement}
Pougu{\'e}-Biyong, J., Semenova, V., Matton, A., Han, R., Kim, A., Lambiotte, R., and Farmer, D.
\newblock Debagreement: A comment-reply dataset for (dis) agreement detection in online debates.
\newblock In \emph{Thirty-fifth conference on neural information processing systems datasets and benchmarks track (round 2)}, 2021.

\bibitem[Shamsi et~al.(2022)Shamsi, Victor, Kantarcioglu, Gel, and Akcora]{shamsi2022chartalist}
Shamsi, K., Victor, F., Kantarcioglu, M., Gel, Y., and Akcora, C.~G.
\newblock Chartalist: Labeled graph datasets for utxo and account-based blockchains.
\newblock \emph{Advances in Neural Information Processing Systems}, 35:\penalty0 34926--34939, 2022.

\bibitem[Wang \& Kim(2018)Wang and Kim]{wang2018predicting}
Wang, J. and Kim, J.
\newblock Predicting stock price trend using macd optimized by historical volatility.
\newblock \emph{Mathematical Problems in Engineering}, 2018:\penalty0 1--12, 2018.

\bibitem[Wang et~al.(2022)Wang, Chen, Xu, Wu, Shen, Xuan, and Yang]{wang2022tsgn}
Wang, J., Chen, P., Xu, X., Wu, J., Shen, M., Xuan, Q., and Yang, X.
\newblock Tsgn: Transaction subgraph networks assisting phishing detection in ethereum.
\newblock \emph{arXiv preprint arXiv:2208.12938}, 2022.

\bibitem[Wang et~al.(2015)Wang, Tang, JIANG, Chen, Wang, and He]{wang2025agenttaxo}
Wang, Q., Tang, Z., JIANG, Z., Chen, N., Wang, T., and He, B.
\newblock Agenttaxo: Dissecting and benchmarking token distribution of llm multi-agent systems.
\newblock In \emph{ICLR 2025 Workshop on Foundation Models in the Wild}, 2015.

\bibitem[Wang et~al.(2023)Wang, Zhang, Liu, Lu, Luo, and He]{wang2023ex}
Wang, Q., Zhang, Z., Liu, Z., Lu, S., Luo, B., and He, B.
\newblock Ex-graph: A pioneering dataset bridging ethereum and x.
\newblock \emph{arXiv preprint arXiv:2310.01015}, 2023.

\bibitem[Wang et~al.(2024)Wang, Wang, Li, Liang, and He]{wang2024megaagent}
Wang, Q., Wang, T., Li, Q., Liang, J., and He, B.
\newblock Megaagent: A practical framework for autonomous cooperation in large-scale llm agent systems.
\newblock \emph{arXiv preprint arXiv:2408.09955}, 2024.

\bibitem[Wang et~al.(2025{\natexlab{a}})Wang, Lou, Tang, Chen, Zhao, Zhang, Song, and He]{wang2025assessing}
Wang, Q., Lou, Z., Tang, Z., Chen, N., Zhao, X., Zhang, W., Song, D., and He, B.
\newblock Assessing judging bias in large reasoning models: An empirical study.
\newblock \emph{arXiv preprint arXiv:2504.09946}, 2025{\natexlab{a}}.

\bibitem[Wang et~al.(2025{\natexlab{b}})Wang, Wu, Tang, Luo, Chen, Chen, and He]{wang2025limits}
Wang, Q., Wu, J., Tang, Z., Luo, B., Chen, N., Chen, W., and He, B.
\newblock What limits llm-based human simulation: Llms or our design?
\newblock \emph{arXiv preprint arXiv:2501.08579}, 2025{\natexlab{b}}.

\bibitem[Wu et~al.(2023)Wu, Irsoy, Lu, Dabravolski, Dredze, Gehrmann, Kambadur, Rosenberg, and Mann]{wu2023bloomberggpt}
Wu, S., Irsoy, O., Lu, S., Dabravolski, V., Dredze, M., Gehrmann, S., Kambadur, P., Rosenberg, D., and Mann, G.
\newblock Bloomberggpt: A large language model for finance.
\newblock \emph{arXiv preprint arXiv:2303.17564}, 2023.

\bibitem[Xie et~al.(2023)Xie, Han, Zhang, Lai, Peng, Lopez-Lira, and Huang]{xie2023pixiu}
Xie, Q., Han, W., Zhang, X., Lai, Y., Peng, M., Lopez-Lira, A., and Huang, J.
\newblock Pixiu: A large language model, instruction data and evaluation benchmark for finance.
\newblock \emph{arXiv preprint arXiv:2306.05443}, 2023.

\bibitem[Yu et~al.(2024)Yu, Li, Chen, Jiang, Li, Zhang, Liu, Suchow, and Khashanah]{yu2024finmem}
Yu, Y., Li, H., Chen, Z., Jiang, Y., Li, Y., Zhang, D., Liu, R., Suchow, J.~W., and Khashanah, K.
\newblock Finmem: A performance-enhanced llm trading agent with layered memory and character design.
\newblock In \emph{Proceedings of the AAAI Symposium Series}, volume~3, pp.\  595--597, 2024.

\bibitem[Zhang et~al.(2024{\natexlab{a}})Zhang, Zhao, Xia, Sun, Sun, Qin, Li, Zhao, Zhao, Cai, et~al.]{zhang2024multimodal}
Zhang, W., Zhao, L., Xia, H., Sun, S., Sun, J., Qin, M., Li, X., Zhao, Y., Zhao, Y., Cai, X., et~al.
\newblock A multimodal foundation agent for financial trading: Tool-augmented, diversified, and generalist.
\newblock In \emph{Proceedings of the 30th ACM SIGKDD Conference on Knowledge Discovery and Data Mining}, pp.\  4314--4325, 2024{\natexlab{a}}.

\bibitem[Zhang et~al.(2024{\natexlab{b}})Zhang, Luo, Lu, and He]{zhang2024live}
Zhang, Z., Luo, B., Lu, S., and He, B.
\newblock Live graph lab: Towards open, dynamic and real transaction graphs with nft.
\newblock \emph{Advances in Neural Information Processing Systems}, 36, 2024{\natexlab{b}}.

\end{thebibliography}
\bibliographystyle{icml2025}

\newpage
\appendix
\onecolumn
\section*{Appendix}

\section{Baseline Strategies} \label{baselines}

To benchmark the performance of MAS, we compare it against a comprehensive set of widely recognized baseline trading strategies. These strategies include traditional rule-based technical indicators and learning-based approaches. All baseline strategies are evaluated across multiple time resolutions (5-minute, 15-minute, 1-hour and 4-hour) to assess performance under different temporal granularities.

\begin{enumerate}
    \item \textbf{SMA \cite{gencay1996non}:} The Simple Moving Average (SMA) strategy makes buy and sell decisions by comparing the asset's price to its average over a specified period. We experiment with different time windows $[5, 10, 15, 20, 25, 30]$, selecting the period that performs best on a validation dataset.
    
    \item \textbf{MACD \cite{wang2018predicting}:} The Moving Average Convergence Divergence (MACD) strategy identifies buy and sell signals by analyzing momentum shifts. It calculates the difference between a 12-day and a 26-day Exponential Moving Average (EMA), with a 9-day EMA acting as a trigger line. EMAs assign greater significance to recent data points.

    \item \textbf{LSTM \cite{ferdiansyah2019lstm}:} This strategy involves comparing today's price with the predicted price for tomorrow to identify potential buying and selling opportunities. If the predicted next price is higher than the current, the model buys; if lower, it sells. This serves as a learning-based baseline and adapts naturally across different temporal resolutions. We choose the same parameters used in CryptoTrade \cite{li2024reflective}.

    \item \textbf{CryptoTrade \cite{li2024reflective}:} This strategy is an LLM-based trading agent designed specifically for cryptocurrency markets, expanding the typical application of LLMs beyond stock market trading. Experiments show that CryptoTrade outperforms time-series baselines in maximizing returns, though traditional trading signals still perform better under most of conditions.
\end{enumerate}

\section{Prompts} \label{prompts}
In this section, we provide the prompts we used for the trading agents.

\begin{tcolorbox}[title=Example Prompt Set for Multi-Agent Trading (Template), colback=purple!5!white,colframe=purple!75!black, fonttitle=\bfseries]
\footnotesize
\textbf{Price Analysis Prompt:}

You are a \{COIN\} cryptocurrency trading analyst. The recent price and auxiliary information is given in chronological order below:

\textless Chronological list of recent prices and technical indicators\textgreater

Write one concise paragraph to analyze the recent information and estimate the market trend accordingly.

\vspace{0.5em}

\textbf{News Analysis Prompt:}

You are a \{COIN\} cryptocurrency trading analyst. You are required to analyze the following news articles:

\textless List of recent news headlines\textgreater

Write one concise paragraph to analyze the news and estimate the market trend accordingly.

\vspace{0.5em}

\textbf{Reflection Prompt:}

You are a \{COIN\} cryptocurrency trading analyst. Your analysis and action history is given in chronological order:

\textless Reasoning, actions, and daily returns from recent steps\textgreater

Reflect on your recent performance and instruct your future trades from a high level, e.g., identify what information is currently more important, and what to be next, like aggressive or conservative. Write one concise paragraph to reflect on your recent trading performance with a focus on the effective strategies and information that led to the most successful outcomes, and the ineffective strategies and information that led to loss of profit. Identify key trends and indicators in the current cryptocurrency market that are likely to influence future trades. Also assess whether a more aggressive or conservative trading approach is warranted.

\vspace{0.5em}

\textbf{Final Decision Prompt:}

You are an experienced \{COIN\} cryptocurrency trader and you are trying to maximize your overall profit by trading \{COIN\}. In each day, you will make an action to buy or sell \{COIN\}. You are assisted by a few analysts below and need to decide the final action.

ON-CHAIN ANALYST REPORT: \\
\textless Price Analysis Output\textgreater

NEWS ANALYST REPORT: \\
\textless News Analysis Output\textgreater

REFLECTION ANALYST REPORT: \\
\textless Reflection Output\textgreater

Now, start your response with your brief reasoning over the given reports. Then, based on the synthesized reports, conclude a clear market trend, emphasizing long-term strategies over short-term gains. Finally, indicate your trading action as a 1-decimal float in the range of [-1,1], reflecting your confidence in the market trend and your strategic decision to manage risk appropriately.
\end{tcolorbox}

\begin{tcolorbox}[title=Example News Analyst Reasoning for BTC (Real Trial), colback=gray!2!white, colframe=gray!70!black, fonttitle=\bfseries]
\footnotesize

\textbf{NEWS ANALYST PROMPT:}

You are a BTC cryptocurrency trading analyst. You are required to analyze the following news articles:

\textit{best way to buy large amount of bitcoin:} \\
i would like to add bitcoin to my asset allocation. would buy a 7 figure amount. all my liquid asset...

\vspace{0.4em}
\textit{How hard would it be to manipulate btc for profit?:} \\
If a whale bought a $\sim$30 billion worth of Bitcoin would it not skyrocket the price?...

\vspace{0.4em}
\textit{Why is anyone talking about segwit these days?:} \\
As the title states, why don't we hear about segwit?...

Write one concise paragraph to analyze the news and estimate the market trend accordingly.

\textbf{NEWS ANALYST RESPONSE:}

The news articles indicate heightened interest in large-scale Bitcoin investments and market dynamics, suggesting potential bullish sentiment. A user's intent to allocate a seven-figure sum to Bitcoin reflects institutional or high-net-worth investor confidence, which could drive demand. Meanwhile, discussions about market manipulation via whale activity highlight concerns over volatility but also underscore Bitcoin's liquidity and sensitivity to large inflows. Technical debates and storage strategies suggest a mature market focused on scalability and security. Overall, sentiment leans positive, with increased buying pressure likely to push prices upward, though volatility risks remain due to potential whale actions.

\end{tcolorbox}

    
    
    
    
    

    

\section{Data Ethics} \label{data-ethics}

We obtain Reddit discussion data via the official Reddit Data API\footnote{\url{https://www.reddit.com/dev/api}}. In accordance with Reddit's Data API Terms\footnote{\url{https://redditinc.com/policies/data-api-terms}}, we are granted a limited, non-exclusive, non-transferable, and non-sublicensable license to access and use Reddit's content and services solely for non-commercial, research, and personal purposes. We strictly refrain from any commercial use or redistribution of Reddit content, and fully comply with all applicable terms and policies. All data collection and usage in this study adhere to the guidelines and restrictions set forth by Reddit Inc.

\section{Parameter sensitivity analysis} \label{para_selec}

\begin{figure}[ht]
    \centering
    \begin{minipage}[t]{0.48\textwidth}
        \centering
        \includegraphics[width=\linewidth]{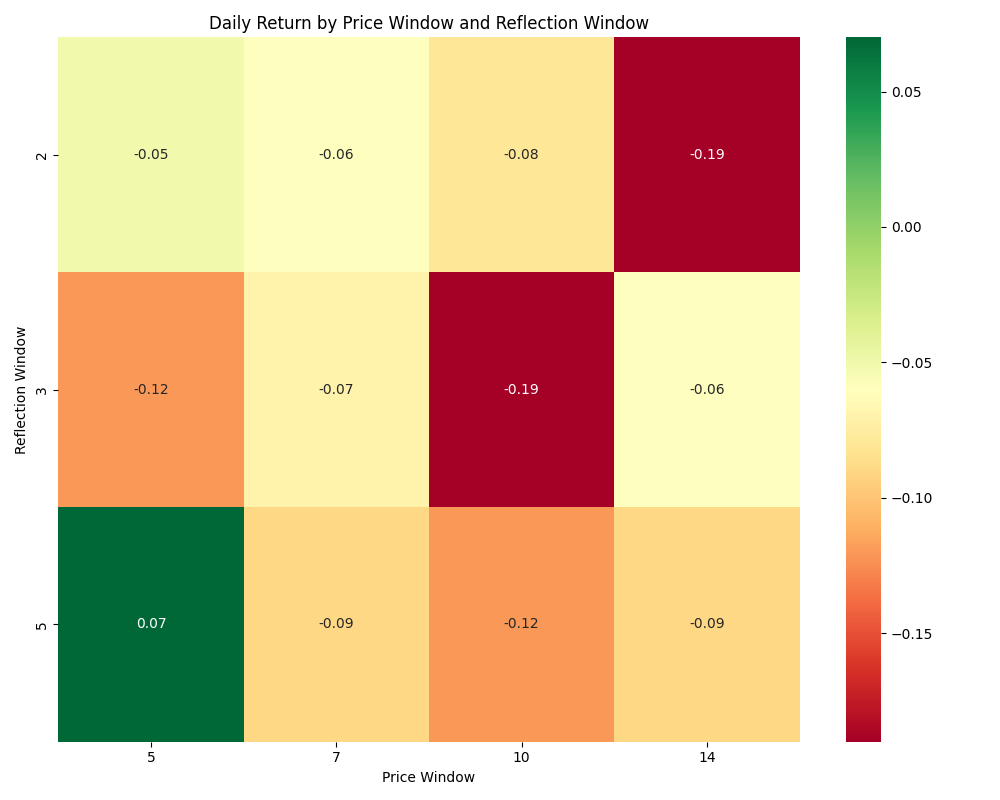}
        {(a) Heatmap of Daily Return}
    \end{minipage}
    \hfill
    \begin{minipage}[t]{0.48\textwidth}
        \centering
        \includegraphics[width=\linewidth]{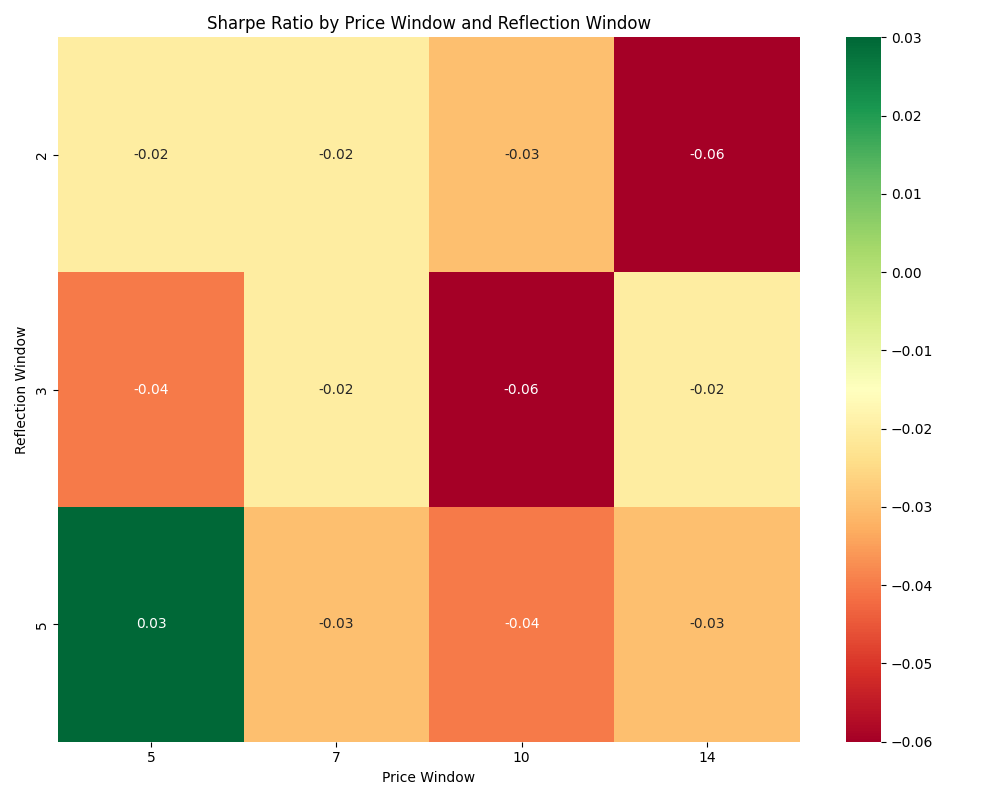}
        {(b) Heatmap of Sharpe Ratio}
    \end{minipage}
    \caption{Parameter sensitivity analysis.}
    \label{fig:heatmaps}
\end{figure}

Figure~\ref{fig:heatmaps} presents a parameter sensitivity analysis for the two key hyperparameters in our framework: the price window and the reflection window. The left panel (a) shows the impact of different parameter combinations on daily return, while the right panel (b) reports the corresponding Sharpe ratio outcomes.

From the heatmaps, we observe that both metrics are highly sensitive to the choice of window sizes. Specifically, larger price and reflection windows tend to result in lower returns and Sharpe ratios, likely due to the reduced responsiveness of the strategy to rapid market changes. Notably, the optimal performance is achieved when the reflection window is set to 5 and the price window to 5, yielding the highest daily return and Sharpe ratio among all tested settings. These results highlight the importance of careful hyperparameter tuning for maximizing trading strategy performance in high-frequency scenarios.

\end{document}